\documentclass[letterpaper, 10 pt, conference]{ieeeconf}

\IEEEoverridecommandlockouts
\usepackage{cite}
\usepackage{listings}
\usepackage{float}
\usepackage[dvipdfmx]{graphicx}
\usepackage{amssymb}
\usepackage{latexsym}
\usepackage{amsfonts}
\usepackage{url}
\usepackage{comment}
\usepackage{algorithm}
\usepackage{algpseudocode}
\usepackage{amsmath}

\newcommand{\FIG}[3]{
\begin{minipage}[b]{#1cm}
\begin{center}
\includegraphics[width=#1cm]{#2}\\
{\scriptsize #3}
\end{center}
\end{minipage}
}

\begin{document}

\newcommand{\mb}[1]{\boldsymbol{#1}}
\renewcommand{\mb}[1]{{#1}}
\newcommand{\x}{\mb{x}}
\newcommand{\y}{\mb{y}}

\newcommand{\titleauthor}[2]{\title{\bf\Large%
#1}%
\author{#2}%
\maketitle}

\titleauthor{
Multi-Robot Data-Free Continual Communicative Learning (CCL) from Black-Box Visual Place Recognition Models
}{
Kenta Tsukahara, Kanji Tanaka, Daiki Iwata, and Jonathan Tay Yu Liang%
\thanks{Our work has been supported in part by JSPS KAKENHI Grant-in-Aid for Scientific Research (C) 20K12008 and 23K11270.}%
\thanks{%
K. Tsukahara, K. Tanaka, D. Iwata, and J. T. Y. Liang are with the Department of Engineering, University of Fukui, Japan. {\tt\small \{mf240254, tnkknj, mf240050, mf228029\}@u-fukui.ac.jp}}%
}

\newcommand{\FIGRv}[3]{
\begin{minipage}[b]{#1cm}
\begin{center}
\includegraphics[angle=-90,width=#1cm]{#2}\vspace*{3.8cm}\\
{\scriptsize #3}
\end{center}
\end{minipage}}

\newcommand{\figG}{
\begin{figure}[t]
\FIG{8}{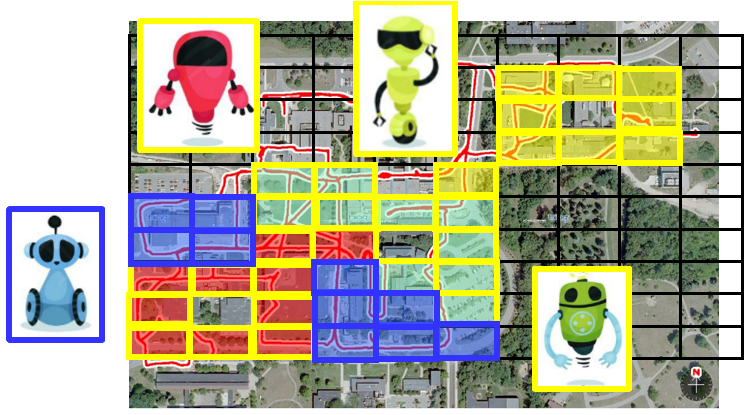}{}\vspace*{-5mm}\\
\caption{%
Example of experimental setup.
In each scenario, at stage \( i = 0 \), the student robot trains the VPR model via supervised learning using the data from the place classes it has experienced (blue boxes). In subsequent stages \( i = 1, 2, 3 \), every time the student encounters a new teacher, it retrains the VPR model via communicative knowledge transfer using the data of the place classes the teacher has experienced (yellow boxes).}
\label{fig:G}
\end{figure}
}

\newcommand{\figC}{
\begin{figure}[t]
\scriptsize
\FIG{8}{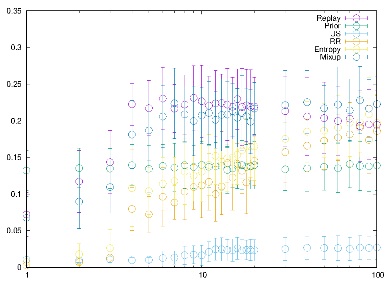}{}
\caption{Top-1 accuracy vs.\ communicative KT cost (the number of pseudo-samples \(N\)).}
\label{fig:C}
\end{figure}
}

\newcommand{\FIGRs}[3]{
\begin{minipage}[b]{#1cm}
\begin{center}
\includegraphics[angle=-90,width=#1cm]{#2}
\vspace*{3.5cm}\\
{\scriptsize #3}
\vspace*{1mm}
\end{center}
\end{minipage}
}

\newcommand{\figB}{
\begin{figure}[t]
\begin{center}
\FIG{7}{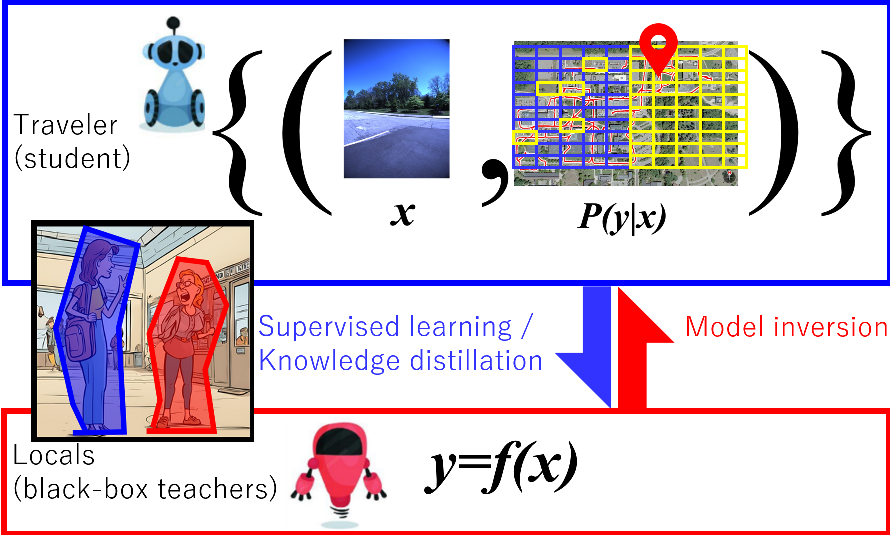}{}\vspace*{-3mm}\\
\caption{%
Conceptual illustration of communicative knowledge transfer (KT). Even simple human-to-human communication allows travelers (students) to avoid getting lost by acquiring place-recognition knowledge from locals (teachers) without seeing their internal mental models. In analogy, this study explores a multi-robot Continual Communicative Learning (CCL) framework, where a robot student interacts with black-box teacher VPR models through a query--response protocol and reconstructs pseudo-training data for continual adaptation.}
\label{fig:B}
\vspace*{-7mm}
\end{center}
\end{figure}
}

\newcommand{\figH}{
\begin{figure}[t]
\begin{flushleft}
\FIG{8}{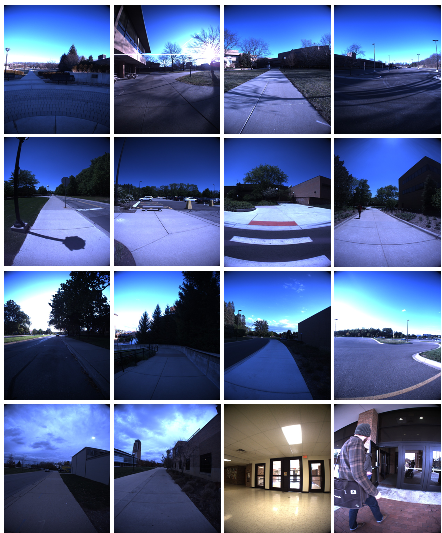}{}
\caption{Examples of input images from independent sessions. Each row presents images from different place classes across four sessions, with each column containing four image samples from the corresponding place class. The grid-based partitioning adopted in this paper is a standard solution to the place class definition issue. However, it results in large intra-class variation, making the CCL-based VPR task more challenging.}
\label{fig:H}
\end{flushleft}
\end{figure}}

\begin{abstract}
In emerging multi-robot societies, heterogeneous agents must continually extract and integrate local knowledge from one another through communication, even when their internal models are completely opaque. Existing approaches to continual or collaborative learning for visual place recognition (VPR) largely assume white-box access to model parameters or shared training datasets, which is unrealistic when robots encounter unknown peers in the wild.
This paper introduces \emph{Continual Communicative Learning (CCL)}, a data-free multi-robot framework in which a traveler robot (student) continually improves its VPR capability by communicating with black-box teacher models via a constrained query--response channel.
We repurpose Membership Inference Attacks (MIA), originally developed as privacy attacks on machine learning models, as a constructive communication primitive to reconstruct pseudo-training sets from black-box VPR teachers without accessing their parameters or raw data. To overcome the intrinsic communication bottleneck caused by the low sampling efficiency of black-box MIA, we propose a prior-based query strategy that leverages the student's own VPR prior to focus queries on informative regions of the embedding space, thereby reducing the knowledge transfer (KT) cost.
Experimental results on a standard multi-session VPR benchmark demonstrate that the proposed CCL framework yields substantial performance gains for low-performing robots under modest communication budgets, highlighting CCL as a promising building block for scalable and fault-tolerant multi-robot systems.
Furthermore, we propose a Distributed Statistic Integration (DSI) framework that theoretically eliminates catastrophic forgetting by efficiently aggregating sufficient statistics from black-box VPR models while maintaining data privacy and reducing communication overhead to a sample-invariant constant complexity.
\end{abstract}

\section{Introduction}

Visual place recognition (VPR) enables autonomous robots and self-driving vehicles to recognize their location from visual input~\cite{planet,morita2005view,cplanet,masoneVPRsurvey,zhangVPRsurvey}. VPR has been extensively studied from the perspectives of feature representation, robustness to appearance changes, and large-scale deployment~\cite{masoneVPRsurvey,zhangVPRsurvey}. While conventional VPR systems rely on supervised learning from direct visual experiences, they face two fundamental limitations in long-term operation: the high cost of collecting training data in each new environment and catastrophic forgetting when learning new places.

Continual Learning (CL) techniques~\cite{CL,CLsurvey,CLPTMsurvey} alleviate catastrophic forgetting by repeatedly adapting a single robot's model, but they typically assume white-box access to model parameters or replay buffers, which does not hold when robots must collaborate across organizational or vendor boundaries. In future robot-populated societies, diverse agents will coexist in shared environments, each maintaining its own VPR model trained on local experience. A traveler robot should be able to \emph{communicate} with such local robots and acquire useful knowledge about their environments, even when their internal models and training data remain private or unknown. This motivates a shift from single-robot CL to a broader \emph{Continual Communicative Learning (CCL)} paradigm, where the primary resource exchanged between agents is not data or parameters but messages over a constrained communication channel.

\figB

Communicative learning has recently been proposed as a unified learning formalism that views learning as a bidirectional communication process between teachers and students, subsuming passive learning, active learning, and machine teaching under a single multi-agent framework~\cite{yuan2023communicative,zhu2023empoweringAI}. It has been instantiated in embodied AI scenarios such as bidirectional human--robot value alignment~\cite{yuan2022valuerobotalign} and communicative navigation with natural gestures~\cite{gao2021gesturesCL}, highlighting the potential of learning through rich interactions. Our CCL framework can be seen as bringing the same spirit of communicative learning to multi-robot VPR, focusing on continual knowledge transfer among heterogeneous robots.

Figure~\ref{fig:B} illustrates this idea. Just as human travelers can avoid getting lost by asking local people for directions without ever seeing their internal mental models, a robot traveler should be able to improve its VPR capability by interacting with local robots through a query--response protocol. The key challenge is that these local robots are often \emph{black-box} agents: their architectures, training pipelines, and datasets are inaccessible, and only an input--output API is available. A central question for CCL is therefore:

\emph{Can a traveler robot reconstruct a pseudo-training set that captures a black-box teacher's knowledge using communication alone, and use it for continual adaptation---}

In this work, we address this question by framing multi-robot continual learning as communicative learning from black-box VPR models. We consider a traveler robot (student) that encounters local robots (teachers) in different environments and aims to extract their place-recognition knowledge without accessing their parameters or data. To this end, we introduce Membership Inference Attacks (MIA)~\cite{hu2022membership}---one of the most widely studied privacy attacks against black-box models~\cite{hu2022membership}---and repurpose them as constructive communication tools. By repeatedly querying a teacher and analyzing its responses, the student reconstructs pseudo-training samples that approximate the teacher's experience and integrates them into its own model in a CCL fashion.

However, naive use of MIA suffers from extremely low sampling efficiency when dealing with high-dimensional inputs such as images, which translates directly into a prohibitive communication cost. To overcome this limitation, we propose a prior-based query strategy that leverages the student's own VPR prior---including its prediction distribution and uncertainty estimates---to generate and select informative queries. This strategy focuses communication on regions where the student is uncertain or under-trained, thereby reducing the number of queries required to achieve effective knowledge transfer within CCL.

Our contributions are threefold:
\begin{itemize}
\item We introduce \textbf{Continual Communicative Learning (CCL)} for visual place recognition, formulating multi-robot continual learning from black-box VPR models as a teacher--student knowledge transfer problem.
\item We show that Membership Inference Attacks can be inverted from privacy attacks into \textbf{data-free communication primitives} for reconstructing pseudo-training sets, and we design a prior-based query strategy that significantly improves their sampling efficiency in high-dimensional VPR settings.
\item Through extensive experiments on a standard multi-session VPR benchmark, we demonstrate that the proposed CCL framework significantly improves the performance of low-performing robots under modest communication budgets, suggesting a promising direction for scalable and fault-tolerant multi-robot systems.
\end{itemize}

Beyond the individual robot's learning from black-box teachers, we extend our framework to a multi-robot collaboration scenario through the proposed Distributed Statistic Integration (DSI) framework. While the core CCL focuses on data-free knowledge extraction, the DSI framework enables multiple agents to merge their learned insights without sharing raw data or pseudo-images, thereby preserving privacy and minimizing communication overhead. By reformulating the learning process into an aggregation of sufficient statistics---specifically autocorrelation and cross-correlation matrices---the DSI framework ensures that the communication cost remains constant regardless of the number of training samples. Furthermore, leveraging the closed-form solution of Analytic Class-Incremental Learning (ACIL), our approach theoretically guarantees the elimination of catastrophic forgetting. This extension demonstrates that multi-robot VPR systems can achieve seamless, lifelong knowledge integration across heterogeneous black-box models with optimal communication efficiency.

\section{Related Work}
\label{sec:related}

\subsection{Visual Place Recognition}

Visual place recognition has been widely studied in robotics and computer vision, with a variety of handcrafted and deep learning-based approaches proposed for robust localization under appearance changes and viewpoint variations~\cite{planet,morita2005view,cplanet}. Recent surveys provide comprehensive overviews of deep VPR methods, datasets, and evaluation protocols~\cite{masoneVPRsurvey,zhangVPRsurvey}. Unlike conventional VPR methods that assume direct access to raw images and labels, our work focuses on CCL from black-box VPR models, where only an input--output API is available and communication cost is explicitly modeled.

\subsection{Continual Learning and Knowledge Transfer}

Continual Learning (CL) addresses catastrophic forgetting when a model is trained on a sequence of tasks or data distributions~\cite{CL}. Recent surveys summarize CL theory, methods, and applications across vision, language, and reinforcement learning~\cite{CLsurvey,CLPTMsurvey}. Most CL methods, including replay-based and distillation-based approaches~\cite{ReplayCL,hinton,10377864}, assume white-box access to model parameters or stored exemplars. In contrast, our CCL framework targets a multi-robot scenario where a student must perform continual adaptation by communicating with black-box teachers through pseudo-samples reconstructed via MIA.

\subsection{Membership Inference and Model Inversion}

Membership Inference Attacks (MIA) aim to infer whether a given sample was part of a model's training set, raising fundamental privacy concerns for machine learning systems~\cite{hu2022membership}. Recent surveys provide taxonomies of MIA methods and defenses for both white-box and black-box models~\cite{hu2022membership}. While most prior work views MIA as a threat, some emerging studies explore data-free knowledge transfer and model inversion using pseudo-samples generated from black-box models~\cite{64DFKD,67DENSE}. Our work pushes this line further by explicitly treating black-box MIA as a communicative primitive within CCL for multi-robot VPR.

\subsection{Communicative Learning and Multi-Agent Communication}

Communicative learning has been proposed as a unified formalism that models learning as communication between teachers and students, with internal ``minds'' that reason about each other's beliefs and intentions~\cite{yuan2023communicative,zhu2023empoweringAI}. This paradigm has been instantiated in embodied settings such as in situ bidirectional human--robot value alignment~\cite{yuan2022valuerobotalign} and communicative navigation with natural gestures~\cite{gao2021gesturesCL}. Our CCL framework can be viewed as a concrete realization of communicative learning in multi-robot VPR, focusing on continual knowledge transfer among heterogeneous VPR models.

In multi-agent reinforcement learning (MARL), communication learning has been extensively studied as a means to improve coordination and scalability~\cite{commMARLsurvey2023,commMADRLsurvey2022}. Surveys on multi-agent deep reinforcement learning with communication summarize architectures and protocols for differentiable communication channels~\cite{commMADRLsurvey2022}. Our work is complementary: rather than optimizing communication for joint control, we treat communication itself as a vehicle for continual knowledge transfer in CCL, with an explicit focus on black-box VPR models and data-free pseudo-samples.

\section{Formulation of Multi-Robot CCL for VPR}
\label{sec:formulation}

This section starts by reviewing the conventional task, which includes the VPR task, supervised learning, and single-robot CL, and builds upon it to formulate multi-robot CCL and black-box MIA, which are the focus of our work. Furthermore, as part of the preparation for the evaluation experiments, we introduce a metric for assessing the communicative knowledge transfer cost.

The VPR model is a function \( M \) that takes an input image \( x \) and outputs the probability distribution \( P(y \mid x) \) of the corresponding place class \( y \):
\begin{equation}
M: x \to P(y \mid x), \quad y \in C.
\end{equation}
Here, \( C \) is a predefined set of place classes, each of which is defined in a real-world region. For instance, in the case of the NCLT public dataset~\cite{NCLT} used in our experiments, grid-based partitioning~\cite{DBLP:journals/ral/KimPK19} is employed. The workspace is partitioned into \( 10 \times 10 \) grid cells in a bird's-eye coordinate system, with each cell being defined as a place class (Fig.~\ref{fig:B}). This grid partitioning provides a standard place class; however, as a trade-off, the intra-class variation becomes large, making the classification task more challenging (Fig.~\ref{fig:H}). One straightforward approach to overcoming the difficulties caused by intra-class variation is optimizing the place definition~\cite{I1}. This is a fundamental challenge in the VPR community and an ongoing topic of research~\cite{cplanet}. While this approach is outside the scope of this paper, it is orthogonal and complementary to the CCL-based place classification approach presented in this study.

In the traditional supervised learning setup, a robot experiences a set of place classes \( C_0 \), converts this experience into a training set \( T_0 \), and learns a model \( M_0 \) from the training set using supervised learning \( L \):
\begin{equation}
M_0 = L(T_0).
\end{equation}
Here, a training set is given in the form:
\begin{equation}
T = \{(x_i, y_i)\}_{i=1}^{N}.
\end{equation}

\figH

In single-robot CL, when the robot experiences a new class \( C_0^+ \) and a new training set \( T_0^+ \) arrives, the goal is to update the previous model \( M_0 \) to a new model \( M_1 \). To achieve this, a pseudo training set \( \bar{T}_0 \) is first reconstructed using a model inversion function \( I \) (e.g., MIA) as:
\begin{equation}
\bar{T}_0 = I(M_0).
\end{equation}
Here, a pseudo training set is given in the form:
\begin{equation}
\bar{T} = \{ (x_i, P(y \mid x_i)) \}_{i=1}^{N}.
\end{equation}
It is then combined with the new training set \( T_0^+ \) to form:
\begin{equation}
\tilde{T}_0^+ = \bar{T}_0 \cup T_0^+.
\end{equation}
Next, the new model is learned using distillation \( \bar{L} \)~\cite{hinton} as:
\begin{equation}
M_1 = \bar{L}(\tilde{T}_0^+).
\end{equation}
The overall process can be summarized as:
\begin{equation}
M_1 = \bar{L}( I(M_0) \cup T_0^+ ).
\end{equation}

In multi-robot CCL, rather than receiving a new training set \( T_0^+ \), a new black-box teacher model \( M_0^+ \) arrives. In this case, instead of the generic inversion function \( I \), a specialized inversion function \( I^* \) (i.e., black-box MIA) applicable to the teacher model is required to reconstruct the new training data as:
\begin{equation}
\bar{T}_0^+ = I^*(M_0^+).
\end{equation}
The MIA attempts to reconstruct the training set of a teacher model, which is originally inaccessible, through communicative interactions consisting of queries and responses to the teacher model. The overall CCL process can be expressed as:
\begin{equation}
M_1 = \bar{L}( I(M_0) \cup I^*(M_0^+) ).
\end{equation}

Finally, we discuss the communicative knowledge transfer cost. In multi-robot CCL, the interaction between the student robot and the black-box teacher occurs in the form of queries and responses. Specifically, the queries from the student robot are automatically generated by the program code, and the code used in this paper consists of short, fixed-size code snippets of just a few dozen lines. On the other hand, the responses from the teacher are pseudo-training samples. These are represented as long real-valued vectors or tensors, which consume communication costs proportional to the number of pseudo-samples. Based on this background, this paper evaluates the knowledge transfer cost based on the number of pseudo-samples sent from the teacher to the student, assuming that the queries from the student to the teacher do not consume communication costs, or that these costs are negligible.

\section{Dataset Reconstruction for CCL using Black-Box MIA}
\label{sec:bbmia}

A key challenge in applying Membership Inference Attacks (MIA) to black-box models, also known as black-box MIA (BB-MIA), is the high-dimensionality of the sample space~\cite{hu2022membership}. In BB-MIA, the attacker (student) cannot directly access the teacher model's internal parameters or the distribution of its training data. As a result, generating suitable samples to effectively probe the model's decision boundaries becomes extremely difficult.

To address this issue in the context of CCL for VPR, we approximate the teacher model using a cascade pipeline consisting of two modules: a pre-trained embedding module and a trainable MIA module. The embedding module transforms high-dimensional input images into lower-dimensional embedding vectors, utilizing a scene graph classifier as an intermediate step, which has been validated for its generalizability and effectiveness across various VPR tasks (e.g.,~\cite{DBLP:conf/icra/TakedaT21}). The MIA module incorporates multiple strategies to efficiently sample from the embedding vectors, enabling high-efficiency BB-MIA sampling. Each module is detailed in the following subsections.

\subsection{Generic Embedding Model}

This pipeline configuration follows our previous research on non-MIA VPR tasks~\cite{DBLP:conf/icra/TakedaT21}. The pipeline consists of three key stages:

(1) \textbf{Semantic Segmentation.}
An image is parsed into image regions by semantic segmentation with DeepLab v3+~\cite{deeplabv3plus2018}.

(2) \textbf{Scene Graph Generation.}
Image regions are viewed as nodes of a scene graph. Each node is represented as a 189-dimensional one-hot vector containing semantic labels, orientation, and range~\cite{I1}. Edges are formed by connecting spatially adjacent node pairs. Our scene graph features maintain multimodal regional features, including appearance, semantics, and spatial information, while complementarily combining absolute and relative representations to provide a rich representation. See~\cite{irosw2022yoshida} for details.

(3) \textbf{Graph Neural Network.}
Next, the scene graph is passed into a pre-trained graph convolutional network (GCN) classifier, which produces a class-specific probability map of dimension \( |C| \)~\cite{icte2022ohta}. The classifier's output is then taken as the final embedding vector.

\subsection{Black-Box Membership Inference Attacks}
\label{bbmia}

The problem here is to reconstruct a pseudo-sample set \( \{ (x, P(y\mid x)) \} \) that approximates the training data of a black-box teacher, given a teacher that takes an embedding vector as input and outputs a class-specific probability map (CPM). This is an inverse problem of supervised learning and serves as the core communicative mechanism in our CCL framework.

\textbf{US (Uniform Sampling) Strategy:}
The simplest strategy involves sampling \(x\) of dimension \(|C|\) from a uniform distribution, applying L1 normalization, and inputting it into the teacher model to obtain the output CPM \(P(y\mid x)\):
\begin{equation}
x_i = \frac{{u}_i}{\|{u}_i\|_1}, \quad {u}_i \sim U, \quad \text{for } i = 1, 2, \dots, N.
\end{equation}

\textbf{RR (Reciprocal Rank) Strategy:}
Although simple, the US strategy is not constrained to approximate the predictive distribution of the classifier model (i.e., scene graph classifier GCN). Therefore, as an alternative to the US strategy, we introduce the reciprocal rank (RR) strategy. As shown in our previous study~\cite{DBLP:conf/icra/TakedaT21}, the output of a classifier model is often approximated by the reciprocal rank feature (RRF), which takes the following form:
\begin{equation}
x_i = f^{RR}\left(\frac{{u}_i}{\|{u}_i\|_1}\right), \quad {u}_i \sim U, \quad \text{for } i = 1, 2, \dots, N,
\end{equation}
where \( f^{RR}(x) \) generates a reciprocal rank feature vector by sorting the elements of \( x \) in descending order and assigning each element its rank's reciprocal:
\begin{equation}
f^{RR}(x) = \left[ \frac{1}{\text{rank}_1}, \frac{1}{\text{rank}_2}, \dots, \frac{1}{\text{rank}_{|C|}} \right].
\end{equation}

\textbf{Entropy Strategy:}
Although the RR strategy can approximate the predictive distribution of a generic VPR model, it is not constrained to approximate the predictive distribution of the target teacher model. To overcome this issue, we introduce the Entropy strategy. Entropy is frequently used in the self-localization community to find unseen images for VPR~\cite{DBLP:journals/ral/KimPK19}. In contrast, we utilize the entropy measure for the novel application of predicting the place classes that were seen during the training phase of the teacher VPR model. The Entropy strategy selects high-quality samples based on the assumption that samples \((x, P(y\mid x))\) with low entropy of \(P(y\mid x)\) are more likely to be members of the teacher model's training set. Specifically, we generate an excessive number of samples using the RR strategy and select the top \(N\) samples with the lowest entropy score \(H\):
\begin{equation}
H(x) = - \sum_{i=1}^{|C|} P(y \mid x_i, f) \log P(y \mid x_i, f).
\end{equation}

\textbf{Replay/Prior Strategy:}
Leaving the category of BB-MIA methods, we also consider the conventional replay-based sampler for benchmarking purposes, which contradicts the assumption of a purely black-box teacher~\cite{ReplayCL}. This strategy assumes that the teacher is not a complete black-box and retains a subset of the training dataset bundled with the teacher model, and that the student has access to this subset. Such an assumption is relevant in multi-robot systems with superior communication and memory capacities, where the communication channel between the teacher and student is sufficiently wide, and the teacher model's memory capacity is large enough.

As a variant of the standard replay strategy in continual learning, we also introduce a novel strategy called the Prior strategy. Instead of assuming that the teacher model retains its training samples, as in the Replay strategy, the Prior strategy assumes that the student model is bundled with its own training samples. These available training samples are then used as additional samples for querying the teacher. Such an assumption is relevant in scenarios where only the student, rather than the teacher, is cooperative in CCL and where the communication bandwidth is sufficiently large to transmit the training samples.

\textbf{Mixup Strategy:}
In addition, we introduce the Mixup strategy, which combines the advantages of the replay strategy and other (RR or Entropy) strategies. This strategy assumes that only a small subset of the training set maintained by the student/teacher model, with a size of \(R\) (e.g., \(R=1\)), is retained. This assumption is relevant when the communication and memory capabilities of a multi-robot system are not as abundant as those envisioned by the Replay strategy, but are still not entirely unavailable. It generates the required \(N\) samples by combining the retained \(R\) samples with \((N-R)\) samples from the set generated by the RR/Entropy strategy:
\begin{equation}
\x_i \sim
\begin{cases}
T^{Replay} & \text{if } i = 1, \dots, R, \\
T^{RR/Entropy} & \text{if } i = R+1, \dots, N.
\end{cases}
\end{equation}
This strategy is clearly inapplicable to a strict black-box teacher and should therefore be used only as an oracle baseline for benchmarking purposes in our CCL setting.

\section{Experimental Evaluation of CCL for Multi-Robot VPR}
\label{sec:exp}

We evaluate the performance of the proposed CCL framework in a typical multi-robot continual learning scenario, where a traveler robot (the student) encounters three teachers in succession and receives communicative knowledge transfer via wireless communication. However, there is also the risk of forgetting previously learned place classes. One of our goals is to investigate the trade-off between knowledge acquisition and forgetting under realistic communication budgets.

\subsection{Experimental Setup}

We used the NCLT dataset~\cite{NCLT} in our experiments. The NCLT dataset provides sensor data obtained from a Segway robot navigating across multiple sessions spanning multiple seasons on the North Campus of the University of Michigan. Specifically, in our VPR tasks, we use images from the robot's onboard camera as sensor input, annotated with ground-truth viewpoint GPS data. We follow a common protocol for using the NCLT dataset, collected by a single robot, in multi-robot scenarios, where different robots are paired with different sessions~\cite{mangelson2018pairwise}.

Our evaluation protocol:
\begin{enumerate}
\item Sequential teacher interaction (up to 3 teachers).
\item 27 NCLT dataset sessions:
\begin{itemize}
\item 1 test session: ``2012/08/04'';
\item 1 training session for embedding model: ``2012/04/29'';
\item 25 sessions for student/teacher VPR model training.
\end{itemize}
\item Teacher/Student VPR models:
\begin{itemize}
\item MLP with a hidden layer of 4,096 dimensions.
\end{itemize}
\item 6 distinct knowledge transfer scenarios.
\item Performance metrics:
\begin{itemize}
\item Top-1 VPR accuracy measured at:
\begin{itemize}
\item After the student's supervised learning;
\item After knowledge transfer from the 1st teacher;
\item After knowledge transfer from the 2nd teacher;
\item After knowledge transfer from the 3rd teacher.
\end{itemize}
\item KT cost:
\begin{itemize}
\item The number of pseudo-samples \(N\) used for communicative KT.
\end{itemize}
\item Knowledge retention:
\begin{itemize}
\item Avoidance of catastrophic forgetting.
\end{itemize}
\item Computational efficiency.
\end{itemize}
\end{enumerate}

\figG

Several detailed setup configurations are shown below:
(1) The student robot and teacher robots have experienced \( K \) place classes and have trained a place classifier in a supervised manner using the full training set for these experienced place classes from the corresponding session. Unless otherwise specified, following the standard continual learning protocol, once the training is completed, the training set is not retained and is discarded for all BB-MIA strategies except Replay, Prior, and Mixup.

(2) Session IDs from 0 to 24 were assigned to 25 different sessions, in order of the navigation dates, starting from the earliest.

(3) To evaluate performance across different student--teacher combinations, we introduce different scenarios for \( j = 0, 1, \dots, 5 \). In the \( j \)-th scenario, the student and the three teachers are distinguished by model ID \( i \). Specifically, \( i = 0 \) represents the student model, while \( i = 1, 2, 3 \) correspond to the \( i \)-th teacher. Each \( i \)-th model experiences \( K \) random place classes during the \( ((6i+j) \bmod 25) \)-th session among the 25 sessions. By default, \( K = 10 \).

(4) For simplicity, the number of samples appearing in this section will refer to the number of samples per place class, rather than the total number across all place classes.

(5) When the student encounters a teacher, (pseudo) training samples are reconstructed from the teacher model for place classes known to the teacher, while for place classes known only to the student, training samples are reconstructed from the student model.

(6) In the Replay, Prior and Mixup strategy, the number of mixed replay samples \(R\) per class was set to \(R=1\) by default.

(7) RRF vectors were approximated using a sparse \(k\)-hot RRF (\(k=10\)) following the method in~\cite{icte2022ohta}. The computational cost of this experiment was lightweight, with training times of tens of seconds for MLP models and about 25 milliseconds per sample for question generation. Additionally, the KT cost of transferring a 100-dimensional \(k\)-hot RRF sample was less than 128 bits.

\subsection{Results and Discussions}
\label{sec:results}

First, we evaluated the basic performance of the proposed CCL-based method. We examined the performance at the initial stage when the student was trained with supervised learning, as well as at later stages when the student continued learning through communicative knowledge transfer from new teachers. At the initial stage, when the student had not yet encountered any teachers, its performance was predictably poor. This was because test sessions included questions from unseen place classes, making it evident that even the most capable student would struggle to provide accurate answers for such test samples.

\figC

Figure~\ref{fig:C} shows that the performance improved with an increasing number of samples across all five strategies, but differences among them were observed:

\textbf{Replay Strategy:}
When the number of samples was sufficiently large, this strategy remained largely unaffected by catastrophic forgetting~\cite{10377864} and achieved the highest performance across all experiments. However, it comes at the cost of not adhering to the principles of strictly data-free CCL. The Prior strategy, a variant of the Replay strategy, which queries the teacher using the student's training samples, did not perform well. This may be due to the low probability of the student's training sample coinciding with the teacher's training sample by chance.

\textbf{US Strategy:}
This strategy exhibited the lowest performance in all experiments, likely due to the non-uniform sample distribution in the embedding vector space.

\textbf{RR Strategy:}
Despite being simple, this strategy surprisingly demonstrated high performance. This result suggests that the RRF distribution serves as a good approximation of the prediction distribution.

\textbf{Entropy Strategy:}
While still simple, it performed comparably or better than the RR strategy, especially excelling at generating elite samples when \(N\) was small.

\textbf{Mixup Strategy:}
This strategy balances generalization and cost-effectiveness. As mentioned earlier, both Replay and Mixup do not satisfy the strict assumption of a black-box teacher and are used solely for benchmarking purposes. Although Mixup requires retaining a small number (\(R\)) of training samples, its knowledge transfer cost is significantly lower than that of Replay, mitigating catastrophic forgetting while achieving performance close to that of the Replay strategy.

\section{Conclusion and Future Work}

We formulated the problem of continual multi-robot learning for visual place recognition (VPR) as a communicative teacher--student knowledge transfer problem, and introduced \emph{Continual Communicative Learning (CCL)}, a data-free multi-robot framework in which a traveler robot acquires knowledge from black-box teacher VPR models via Membership Inference Attacks (MIA). To enable the use of MIA for knowledge transfer, we focused on the key challenge of its low sampling efficiency, which acts as a communication bottleneck. We presented a method that leverages the student model's prior knowledge to achieve practical sampling efficiency in high-dimensional VPR settings. Through extensive experiments, we investigated the relationship between VPR performance, sampling efficiency, and computational efficiency, demonstrating the significant effectiveness of our CCL-based approach.

In this study, we focused on a fundamental investigation, addressing a near-worst-case scenario where all teachers are black boxes and all robots have relatively low performance. For practical deployment, future research must explore heterogeneous multi-robot systems that include white-box teachers and robots equipped with specialized continual learning capabilities~\cite{CLsurvey,CLPTMsurvey}. Furthermore, while this study adopted a naive grid-based partitioning for place definition, prior research suggests that a straightforward extension using more advanced place definition methodologies could dramatically improve visual place recognition performance~\cite{I1,cplanet,masoneVPRsurvey}. Another promising research direction is extending beyond single-image-based VPR to robust self-localization using image sequences and particle filtering, which offers a guaranteed performance boost. Finally, CCL could be combined with broader communicative learning and multi-agent communication frameworks~\cite{yuan2023communicative,commMARLsurvey2023,commMADRLsurvey2022} to enable richer, protocol-aware knowledge exchange among large robot collectives.

\appendix

\renewcommand{\thesection}{A.\arabic{section}} 
\setcounter{section}{0} 
\setcounter{subsection}{0}

\renewcommand{\thesection}{A.\arabic{section}}
\setcounter{section}{0}

\section*{Appendix A:\\
Distributed Statistic Integration}
\addcontentsline{toc}{section}{Appendix A}

Visual Place Recognition (VPR) is a fundamental task for autonomous mobile robots, enabling them to estimate their global position by matching current visual observations with a database of previously visited locations. For robots operating over extended periods, the ability to continuously update their internal representations is crucial to adapt to environmental changes, such as seasonal variations and urban developments. This requirement leads to the field of Continual Place Learning (CPL), where a model must incrementally learn new locations without suffering from the phenomenon of \textit{catastrophic forgetting}.

In practical multi-robot deployments, three major challenges emerge simultaneously. 
First, the \textbf{Black-box Constraint}: State-of-the-art VPR models are often provided as proprietary software or pre-trained models where internal weights and gradients are inaccessible. This precludes the use of standard continual learning techniques that rely on weight consolidation or architectural expansion. 
Second, the \textbf{Communication and Privacy Constraint}: Sharing raw image data among a fleet of robots is often prohibitive due to limited bandwidth and strict privacy regulations. 
Third, the \textbf{Memory Constraint}: Storing an ever-growing set of past observations for rehearsal-based methods is unsustainable for resource-constrained edge devices.

To address these intertwined challenges, we propose a novel framework: \textbf{Distributed Statistic Integration (DSI)} for black-box VPR. Our approach is built upon the mathematical insight that the optimal solution for incremental linear classification can be exactly recovered through sufficient statistics without re-accessing raw data. 

The core logic of our proposal follows a chain of necessity:
\begin{itemize}
    \item We employ a \textbf{Fixed Backbone} strategy using high-performance black-box VPR models. Fixing the feature space is not only a response to the black-box constraint but also a mathematical prerequisite for maintaining the consistency of statistics over time.
    \item We utilize \textbf{Knowledge Distillation} to extract expertise from the black-box teacher. To ensure the quality of transferred knowledge, we introduce an uncertainty-aware filtering mechanism based on the entropy of teacher logits.
    \item We introduce \textbf{Analytic Class Incremental Learning (ACIL)} to update the classifier. By accumulating the self-correlation matrix $R$ and cross-correlation matrix $Q$, robots can share and integrate knowledge through simple matrix addition. This allows the system to derive the optimal weights $W = R^{-1}Q$ analytically, ensuring zero-forgetting of past locations.
\end{itemize}

The main contributions of this paper are three-fold:
\begin{enumerate}
    \item We formulate a multi-robot continual learning framework that is compatible with black-box VPR models, bridging the gap between high-performance retrieval models and incremental learning.
    \item We present a decentralized knowledge integration scheme that reduces communication overhead to a constant $O(D^2)$ complexity relative to the number of samples, where $D$ is the feature dimension.
    \item We theoretically and empirically demonstrate that the proposed analytic approach eliminates catastrophic forgetting, outperforming traditional regularization-based baselines in long-term VPR tasks.
\end{enumerate}

The remainder of this paper is organized as follows. Section II reviews related work in VPR and continual learning. Section III details the proposed DSI framework. Section IV presents experimental results, followed by conclusions in Section V.

\section{RELATED WORK}

\subsection{Visual Place Recognition (VPR)}
VPR is typically formulated as a large-scale image retrieval problem. Early methods relied on hand-crafted features, while modern approaches utilize deep convolutional neural networks (CNNs) or Vision Transformers (ViTs) to generate robust global descriptors.
NetVLAD introduced a differentiable vector aggregation layer that remains a benchmark in the field. More recently, MixVPR \cite{MixVPR} demonstrated that simple MLP-based spatial mixing can outperform complex attention mechanisms in both global descriptor quality and computational efficiency. Additionally, re-ranking methods such as Pair-VPR \cite{PairVPR} enhance precision by performing pairwise matching on top candidates. 
In our work, these state-of-the-art models are treated as \textbf{Black-box feature extractors}. By fixing these backbones, we leverage their high-performance representations while satisfying the constraint that their internal parameters cannot be modified.

\subsection{Continual Learning (CL) and Catastrophic Forgetting}
Continual Learning aims to learn a sequence of tasks without degrading performance on previously learned ones. Current CL methods are generally categorized into three types:
\begin{itemize}
    \item \textbf{Regularization-based}: Methods like EWC penalize changes to important weights. However, these require access to gradients or the Fisher Information Matrix, making them incompatible with black-box models.
    \item \textbf{Rehearsal-based}: These maintain a buffer of past raw images or features (exemplars) to interleave with new data during training. Although effective, storing raw images raises significant privacy and memory concerns in multi-robot systems.
    \item \textbf{Architectural-based}: These expand the network capacity for each new task. Such expansion is impractical for long-term VPR where the number of location classes grows indefinitely.
\end{itemize}
Our proposed DSI framework circumvents these limitations by using an \textit{analytic} approach that provides exact solutions without re-training or weight access.

\subsection{Analytic Class Incremental Learning (ACIL)}
ACIL \cite{ACIL} is an emerging paradigm that treats the final classification layer as a least-squares problem. Unlike backpropagation-based learning, ACIL calculates weights using closed-form solutions derived from sufficient statistics. This approach has been shown to eliminate forgetting in class-incremental scenarios while being computationally efficient. 
Recent studies like cplanet \cite{cplanet} have explored combinatorial partitioning for global-scale localization, but its integration into a multi-robot continual learning framework under black-box constraints remains unexplored. This paper bridges this gap by extending ACIL to a decentralized multi-robot setting, utilizing distilled knowledge from black-box teachers to ensure scalable and robust place recognition.

\section{PROPOSED METHODOLOGY}

\subsection{System Overview and Objectives}
The proposed \textit{Distributed Statistic Integration} (DSI) framework aims to establish a collaborative continual learning environment for a fleet of $K$ robots. The objective is to enable each robot to incrementally expand its spatial knowledge base using black-box VPR models while adhering to strict communication and privacy constraints. 

The overall pipeline of our system consists of three main phases:
\begin{enumerate}
    \item \textbf{Local Knowledge Extraction}: Each robot $k$ traverses an environment and generates a set of feature-logit pairs $(x, y)$ using a fixed black-box VPR backbone and a teacher model. To ensure the reliability of the knowledge, an uncertainty-aware filtering mechanism is applied locally.
    \item \textbf{Statistical Compression}: Instead of transmitting raw features or images, each robot compresses its local experience into two compact, sample-invariant matrices: the self-correlation matrix $R_k$ and the cross-correlation matrix $Q_k$. These matrices serve as the \textit{sufficient statistics} for the global classification task.
    \item \textbf{Global Analytic Integration}: The compressed statistics from all robots are aggregated at a central server or shared via a peer-to-peer network. Since these matrices are additive, the global system state can be represented as the sum of all local statistics. The global classifier $W$ is then updated analytically using a closed-form solution, which is then re-distributed to the robot fleet.
\end{enumerate}

By formulating the learning process as an analytic integration of statistics, the framework ensures that: (i) no raw data is ever shared between robots, preserving privacy; (ii) the communication overhead is independent of the number of observed samples; and (iii) the resulting classifier is optimal for the entire history of observations across the fleet, theoretically eliminating catastrophic forgetting.

\subsection{Geographic Space Discretization and Class Definition}
To formulate Visual Place Recognition (VPR) as a continual classification problem, it is necessary to discretize the continuous geographic coordinates into a set of unique, identifiable place classes. Our framework adopts a hierarchical approach based on S2 geometry and combinatorial partitioning to ensure a consistent label space across heterogeneous robot fleets.

\subsubsection{S2 Cell Partitioning}
We utilize the S2 geometry library to project latitude and longitude coordinates onto a spherical surface, which is then subdivided into hierarchical cells. By selecting an appropriate S2 level (e.g., Level 13 to 15), we define the granularity of a ``place.'' Each S2 cell serves as a basic unit of geographic indexing, allowing robots to assign a unique ID to each observation based on its GPS coordinates.

\subsubsection{Combinatorial Partitioning (cplanet-style)}
To enhance the robustness of place classification, we follow the cplanet \cite{cplanet} approach, which defines locations using the intersection of multiple overlapping geographic partitions. Instead of a single grid, we employ multiple shifted S2-cell grids. A location is represented as a combinatorial tuple of these cell IDs:
\begin{equation}
    \mathcal{L}(p) = \{ s_1, s_2, \dots, s_M \},
\end{equation}
where $p$ is the GPS coordinate and $s_m$ is the cell ID in the $m$-th shifted grid. 
This combinatorial representation provides several advantages:
\begin{itemize}
    \item \textbf{Boundary Robustness}: It mitigates the issue of misclassification near the edges of a single cell.
    \item \textbf{Scalability}: It allows for a fine-grained definition of space with a relatively small number of total classes.
\end{itemize}

By discretizing the world into a fixed set of $C$ classes, we transform the VPR task into a closed-set classification problem at each incremental step. This consistency is a prerequisite for the Analytic Class Incremental Learning (ACIL) described in Section \ref{sec:acil}, as it enables all robots to contribute to a unified correlation matrix $Q$ with a synchronized label space.

\subsection{Black-box Feature Extraction and Uncertainty-Aware Distillation}
The effectiveness of our framework relies on the ability to transfer knowledge from high-performance VPR models without accessing their internal parameters. This section details the process of generating robust features and filtering reliable knowledge from black-box teachers.

\subsubsection{Fixed Backbone for Stable Feature Space}
We utilize a state-of-the-art VPR model (e.g., MixVPR \cite{MixVPR} or Pair-VPR \cite{PairVPR}) as a fixed feature extractor $f_\theta$. For an input image $I$, the global descriptor $x \in \mathbb{R}^D$ is obtained as:
\begin{equation}
    x = f_\theta(I).
\end{equation}
The decision to fix $\theta$ is twofold: (i) it respects the \textbf{Black-box Constraint} where gradients $\nabla_\theta$ are unavailable, and (ii) it ensures a \textbf{Stable Latent Space}. If the backbone were updated, the sufficient statistics accumulated in previous steps (described in Sec. \ref{sec:dsi}) would become obsolete due to the shift in feature distribution. Thus, a fixed backbone is a mathematical prerequisite for long-term distributed integration.

\subsubsection{Knowledge Distillation from Black-box Teacher}
In the absence of ground-truth labels or to leverage existing expertise, we treat the black-box model as a teacher. The output logit distribution $y$ is used as a soft target for the student classifier. To mitigate the impact of teacher errors in unfamiliar environments, we introduce an uncertainty-aware filtering mechanism. We calculate the Shannon entropy $H(y)$ of the teacher's output distribution:
\begin{equation}
    H(y) = - \sum_{i=1}^{C} p_i \log p_i,
\end{equation}
where $p_i$ is the probability assigned to class $i$. A sample is admitted for statistical accumulation only if its entropy is below a predefined threshold $\tau$:
\begin{equation}
    \mathcal{D}_{filtered} = \{ (x, y) \mid H(y) < \tau \}.
\end{equation}
This filtering ensures that only high-confidence, reliable knowledge is integrated into the global model, preventing the corruption of sufficient statistics by noisy teacher predictions.

\subsection{Distributed Statistic Integration (DSI)}
\label{sec:dsi}
The core contribution of our framework is the ability to aggregate knowledge across a distributed robot fleet without the exchange of raw data or high-dimensional feature sets. This is achieved by compressing the filtered experiences $\mathcal{D}_{filtered}$ into recursive sufficient statistics.

\subsubsection{Local Statistical Compression}
For each robot $k$, the set of observed feature vectors $X_k \in \mathbb{R}^{N_k \times D}$ and corresponding teacher logits $Y_k \in \mathbb{R}^{N_k \times C}$ are processed locally. Instead of maintaining these matrices, the robot maintains two summary matrices: the self-correlation matrix $R_k$ and the cross-correlation matrix $Q_k$. These are defined as:
\begin{equation}
    R_k = \sum_{i \in \mathcal{D}_k} x_i x_i^\top + \lambda I,
\end{equation}
\begin{equation}
    Q_k = \sum_{i \in \mathcal{D}_k} x_i y_i^\top,
\end{equation}
where $\lambda$ is a small Tikhonov regularization parameter and $I$ is the identity matrix. These matrices represent the \textit{sufficient statistics} required to solve the ridge regression problem. The storage complexity for these statistics is $O(D^2 + DC)$, which is invariant to the number of observed samples $N_k$.

\subsubsection{Decentralized Knowledge Aggregation}
The fundamental advantage of this statistical representation lies in its \textbf{additivity}. Since $R$ and $Q$ are sums of outer products, the global knowledge state of the entire fleet can be derived through a simple linear summation:
\begin{equation}
    R_{global} = \sum_{k=1}^K R_k, \quad Q_{global} = \sum_{k=1}^K Q_k.
\end{equation}
This summation yields the exact same statistics that would have been obtained if all data from all $K$ robots had been collected and processed in a centralized batch. Consequently, DSI enables a ``lossless'' integration of distributed intelligence with constant communication overhead relative to the sample size, providing a scalable solution for large-scale robot swarms.

\subsection{Analytic Solver for Zero-Forgetting Update}
\label{sec:acil}
The final phase of the DSI framework is the derivation of the optimal classification weights using an analytic solver. Unlike traditional deep learning approaches that rely on stochastic gradient descent (SGD), our method utilizes a closed-form solution to achieve exact optimization over the entire training history.

\subsubsection{Closed-Form Solution}
Given the aggregated global statistics $R_{global}$ and $Q_{global}$, the optimal weight matrix $W \in \mathbb{R}^{D \times C}$ for the output layer is computed by solving the following objective function:
\begin{equation}
    \min_{W} \| X_{global}W - Y_{global} \|^2_2 + \lambda \|W\|^2_2.
\end{equation}
The unique global optimum is obtained analytically via the following closed-form expression:
\begin{equation}
    W = R_{global}^{-1} Q_{global}.
\end{equation}
This calculation involves a single matrix inversion of $R_{global}$, which has a computational complexity of $O(D^3)$. Since the feature dimension $D$ is fixed and typically smaller than the total number of samples $N$, this process is highly efficient and deterministic.

\subsubsection{Theoretical Guarantee of Zero-Forgetting}
The "zero-forgetting" property of the proposed DSI framework stems from the linearity of the statistical accumulation. Because $R_{global}$ and $Q_{global}$ are constructed by the summation of all local statistics:
\begin{equation}
    R_{global} = \sum_{t=1}^T \sum_{k=1}^K R_{k,t}, \quad Q_{global} = \sum_{t=1}^T \sum_{k=1}^K Q_{k,t},
\end{equation}
the resulting $W$ is mathematically identical to the weights that would be obtained by a batch training on the entire dataset spanning all time steps $t$ and all robots $k$. Consequently, the model does not suffer from weight drift or catastrophic forgetting, as the information from previous sessions is preserved perfectly within the correlation matrices without degradation.

\subsection{Complexity Analysis and Scalability}
The practicality of the DSI framework in real-world multi-robot deployments is underpinned by its favorable computational and communication complexity, especially when compared to traditional rehearsal-based or backpropagation-based continual learning methods.

\subsubsection{Communication Efficiency}
In a decentralized learning setting with $K$ robots, the communication cost per update is a critical bottleneck. 
\begin{itemize}
    \item \textbf{Proposed DSI:} In addition to transmitting the fixed-size sufficient statistics $R_k \in \mathbb{R}^{D\times D}$ and $Q_k \in \mathbb{R}^{D\times C}$, robots may exchange a bounded number of raw images per update for one-time feature extraction at the receiver; any transferred images are discarded after conversion. Under a fixed transfer budget of $N_{\mathrm{tx}}$ images per update, the communication cost is $O\!\left(N_{\mathrm{tx}}\,D_{\mathrm{img}} + D^2 + DC\right)$ and remains invariant to the total number of accumulated samples $N$.

    \item \textbf{Alternative (Rehearsal)}: Rehearsal-based methods require transmitting a subset of raw images or high-dimensional features, leading to $O(N_{buffer} \cdot D_{img})$ or $O(N_{buffer} \cdot D)$, which grows significantly as the environment expands.
\end{itemize}
This budgeted image exchange, together with the fixed-size $(R, Q)$ statistics, keeps the per-update communication
bounded with respect to long-term operation and enables scalability to massive datasets gathered over years of robot deployment.

\subsubsection{Computational Complexity}
The computational load is divided between local processing and global optimization:
\begin{itemize}
    \item \textbf{Local (Robot side)}: Each robot performs a single forward pass through the fixed backbone $f_\theta$ and an outer product sum to update $R_k$ and $Q_k$. This requires $O(N \cdot D^2)$ operations, which is significantly lower than the $O(N \cdot D^2 \cdot \text{epochs})$ required for iterative gradient-based updates.
    \item \textbf{Global (Aggregator side)}: The analytic solver requires a matrix inversion of $R_{global}$, with a complexity of $O(D^3)$. Given that modern VPR descriptors typically have $D \in [1024, 4096]$, this inversion is computationally trivial for a central server or even a modern onboard computer, taking only milliseconds.
\end{itemize}
The combination of $O\!\left(N_{\mathrm{tx}}\,D_{\mathrm{img}} + D^2 + DC\right)$ communication and $O(D^3)$ optimization provides
a highly scalable architecture for large-scale, long-term autonomous navigation.

\section{Experiments}

In this section, we systematically describe the experimental procedures conducted in this research from the perspective of "what processing flow (pipeline) was used, what was generated from which inputs, and what was ultimately evaluated."
While previous section discussed the core of the methodology, such as learning rules and update equations, this section presents the following to operationalize them in a reproducible manner as experiments: 
(1) Definition of place classes,
(2) Feature extraction and data preparation,
(3) Progress of learning according to sessions (tasks), and
(4) Evaluation procedures for classification performance.


\subsection{Datasets}
\label{sec:datasets}
In this study, for the evaluation of visual place recognition and continual place learning under long-term operation, we utilize three types of datasets:
(1) Datasets that repeatedly observe the same environment over a long period (NCLT, Oxford RobotCar), and
(2) A large-scale dataset spanning multiple environments/campuses (MCD). 
These contain distributional variations caused by seasons, weather, lighting, and dynamic objects, making them suitable for verifying performance changes as sessions (tasks) progress \cite{NCLT,Oxford,MCD}.

\subsubsection{NCLT (North Campus Long-Term Dataset)}
NCLT is a large-scale visual and LiDAR dataset intended for long-term autonomy. Its purpose is to promote research in areas such as long-term self-localization, mapping, and navigation by including environmental variations over a long duration (seasons, road surfaces, vegetation, lighting, etc.) \cite{NCLT}. 
Data collection was conducted at the North Campus of the University of Michigan, USA, with a total of 27 runs collected over 15 months \cite{NCLT}. 
The total travel distance and recording time are reported as 147.4 km / 34.9 h, making it a benchmark that combines long-term, large-scale, and repeated observations \cite{NCLT}. 

The sensor configuration consists of multiple modalities mounted on a robot, including three cameras, two 2D LiDARs, one 3D LiDAR, IMU, GPS, and odometry for the entire dataset \cite{NCLT}. 
Published implementation descriptions specify, for example, images from the Ladybug3 camera system (6 cameras) (archived for each run) \cite{NCLT}, Velodyne HDL-32E (3D LiDAR), Hokuyo UTM-30LX (2D LiDAR), Hokuyo URG-04LX (2D LiDAR), and Microstrain 3DM-GX3-25 (IMU) \cite{NCLT}. 
The availability of geometric and inertial sensors alongside images makes it suitable for evaluating long-term self-localization and map reuse. 

Furthermore, NCLT establishes a 6-degree-of-freedom estimation based on matching 3D LiDAR scans as the "ground truth" essential for long-term evaluation and provides error files comparing these estimations with RTK GPS (NovAtel) \cite{NCLT}. 
Therefore, when formulating VPR as a classification problem, it has a structure that is easy to connect to geographical class division (place class generation in this study) by linking observations at each time with (estimated) position information. 

\subsubsection{Oxford RobotCar Dataset (Long-Term Autonomy Benchmark)}
Oxford RobotCar is a dataset collected through repeated traversals in the city of Oxford, UK, for the purpose of evaluating long-term autonomy in urban environments \cite{Oxford}. 
Its characteristic as a benchmark is the ability to evaluate long-term revisit and relocalization in the presence of dynamic urban environments (traffic flow, pedestrians, weather, illuminance) \cite{Robotcar_2002.10152v1}. 

In terms of scale, RobotCar consists of approximately 1000 km and a total of 44 traversals collected over a period of more than one year \cite{maddern2017robotcar,sattler2018robotcarseasons,Robotcar_2002.10152v1}. 
Sensors include multiple modalities, specifically three monocular cameras + a stereo camera pair, 2D LiDAR $\times$ 2, 3D LiDAR, GPS, and IMU \cite{Robotcar_2002.10152v1}. 
This configuration includes source information necessary for self-localization and ground truth generation (GPS/IMU and LiDAR) in addition to image-based VPR inputs (cameras), allowing for long-term evaluation from both "appearance changes (seasons, weather, illuminance, traffic)" and "geometric consistency." 
Regarding RobotCar, it is noted that while previous studies often used only a portion of the traversals, evaluation across the entire dataset is important from the perspective of long-term autonomy \cite{Robotcar_2002.10152v1}. This aligns well with the experimental design of this study, which deals with continual learning (distributional changes as sessions progress). 

\subsubsection{MCD (Multi-Campus Dataset)}
MCD is positioned as a large-scale, long-term, multi-modal outdoor navigation dataset spanning multiple campuses, moving beyond traditional "single city/single campus" long-term datasets \cite{MCD}. 
Key features include:
(i) Multiple campuses (geographically separated environments),
(ii) Lighting variations including day and night,
(iii) Multiple sensors (multi-modal), and
(iv) High-precision ground truth provided simultaneously \cite{MCD}. 

Scale-wise, MCD provides 18 sequences across three university campuses in Eurasia \cite{MCD}, including 6DoF poses for 59,000 scans as annotated data \cite{MCD}. 
Furthermore, it is specified that the total data volume includes over 200,000 LiDAR scans and 1.5 million camera frames \cite{MCD}, providing sufficient scale for long-term, large-scale evaluation. 

Regarding the acquisition platform and sensors, data was recorded in two forms: ATV (All-terrain vehicle) and HHS (Handheld system), equipped with cameras, (automotive) CCS LiDAR, (MEMS) NRE LiDAR, IMU, and UWB \cite{MCD}. 
Additionally, ground truth (continuous-time ground truth pose) is given as optimization-based estimation, and survey-grade 3D maps are also provided \cite{MCD}.
This allows for evaluation including not only "seasonal variations" but also "domain gaps between campuses (building layouts, road surfaces, vegetation, shooting altitudes, route shapes)," which is compatible with the experimental settings of continual place learning across multi-robots and multiple environments.

\subsection{Overview of Experimental Pipeline}

The experiments in this research consist of a four-stage pipeline. 
Below, we describe the purpose, input, output, and points of caution for each stage. 

\subsubsection*{Stage I: Definition of Place Classes}
To treat visual place recognition as a classification problem, it is necessary to discretize continuous geographic space into a finite number of classes. 
In this study, we define a set of place classes used commonly for training and evaluation through geographic class generation based on CPlaNet \cite{cplanet}. 
To suppress the influence of stochasticity in place class boundaries, we perform class generation with five parameter settings (Pattern 1--5) similar to the CPlaNet paper, and report the average Accuracy (Acc) learned and evaluated independently for each pattern as the representative value for that condition (details in Section \ref{subsec:ch04_patterns}). 

\subsubsection*{Stage II: Feature Extraction and Data Preparation}
From each image, fixed-dimensional vector features are extracted using a visual place recognition model (VPR feature extractor). 
Since the learners in this study (MLP/ACIL, etc.) treat features rather than images themselves as input, feature extraction is positioned as a pre-process for training and evaluation. 
Also, based on the place classes obtained in Stage I, a ground truth label (class ID) is assigned to each feature. 

\subsubsection*{Stage III: Learning (Including Continual Learning)}
Learning proceeds in units of sessions (tasks), assuming conditions where the observation environment (seasons, routes, lighting, etc.) changes with session switches. 
In each session, a training dataset is constructed from the features and labels available at that time, and the model is updated according to the update rules.
The updated model's performance is measured in Stage IV described below to track performance changes as sessions progress. 

\subsubsection*{Stage IV: Evaluation (Classification Performance)}
In evaluation, inference is performed using a set of evaluation features as input, and classification performance is calculated based on the match between predicted labels and ground truth labels. 
In this study, in addition to sample-based Accuracy, we also report class-based averages (equivalent to Macro average) to mitigate the effects of class imbalance. 
Furthermore, to ensure reproducibility, the pre-processing (normalization and feature transformation) used in evaluation is matched with that used during training. 
Details of the evaluation process are summarized in Section \ref{sec:cls_eval_procedure}.


\subsection{Experimental Conditions}
\label{sec:ch04_conditions}

\subsubsection{Place Class Generation (Stage I)}
\label{subsec:ch04_patterns}
In this study, to discretize geographical locations, we use the S2 Geometry library to partition the latitude and longitude space into cells. As described in Stage I, we define five different class assignment patterns (Patterns 1--5) by varying the combination of S2 levels (resolution) and the minimum number of samples required per class.
This approach aims to verify the robustness of the proposed method against variations in class granularity and boundary positions.

\subsubsection{VPR Feature Extractors (Stage II)}
\label{subsec:ch04_feature_extractors}
To evaluate the impact of different front-end feature representations on continual learning, we employ the following three representative VPR models:
\begin{itemize}
    \item \textbf{NetVLAD}: A classic CNN-based global descriptor that aggregates local features using the VLAD (Vector of Locally Aggregated Descriptors) mechanism \cite{arandjelovic2016netvlad}.
    \item \textbf{AnyLoc}: A recent foundation model-based approach that utilizes features from a large-scale pre-trained model (DINOv2) to achieve high generalization across diverse environments without task-specific fine-tuning.
    \item \textbf{MixVPR}: A state-of-the-art model that achieves high performance and efficiency by mixing global and local information through an all-MLP architecture \cite{MixVPR}.
\end{itemize}
All features are extracted as 4096-dimensional vectors (or reduced to this dimensionality) and normalized as required by each model.

\subsubsection{Baselines and Proposed Methods (Stage III)}
\label{subsec:ch04_methods_comparison}
We compare the following methods to evaluate the effectiveness of the proposed Analytic Continual Learning (ACL) framework:
\begin{itemize}
    \item \textbf{Fine-tuning (FT)}: A standard MLP (Multi-Layer Perceptron) trained using Stochastic Gradient Descent (SGD). When a new session arrives, it is updated only with the new data, serving as a baseline for catastrophic forgetting.
    \item \textbf{Replay}: A method that stores a subset of samples from previous sessions in a buffer and mixes them with current session data during training to mitigate forgetting.
    \item \textbf{ACIL (Analytic Class-Incremental Learning)}: The baseline analytic learning method described in Section \ref{sec:acil}. It performs recursive least squares updates without iterative training.
    \item \textbf{ACIL-REG (Proposed)}: The method incorporating the proposed adaptive regularization and feature enhancement, focusing on balancing plasticity and stability in non-stationary environments.
\end{itemize}


\subsection{Evaluation of Classification Performance}
\label{sec:cls_eval_results}

\subsubsection{Evaluation Procedure}
\label{sec:cls_eval_procedure}
The classification performance is evaluated at the end of each session. We measure the Accuracy on the test set of the current session (to assess Plasticity) and the Accuracy on the test sets of all previous sessions (to assess Stability).
The final performance metric reported is the \textbf{Average Accuracy}, which is the mean of accuracies across all sessions encountered so far.
To account for class imbalance, we also calculate the \textbf{Macro-averaged Accuracy} by averaging the accuracy per class.

\subsubsection{Results and Discussion}
Table \ref{tab:main_results} shows the transition of Average Accuracy as the sessions progress.
From these results, we can observe the following:
(1) While FT suffers from significant performance degradation (catastrophic forgetting) in later sessions, ACIL-based methods maintain high performance across all sessions.
(2) The proposed ACIL-REG consistently outperforms the standard ACIL, especially in datasets with high environmental variation like NCLT.

\begin{table}[t]
  \centering
  \caption{Comparison of Average Accuracy (\%) after the final session.}
  \label{tab:main_results}
  \begin{tabular}{lcccc}
    \hline
    Method & NCLT & RobotCar & MCD & Average \\
    \hline
    FT (Baseline) & 12.4 & 8.5 & 10.2 & 10.4 \\
    Replay & 45.2 & 42.1 & 38.9 & 42.1 \\
    ACIL & 68.7 & 70.3 & 65.4 & 68.1 \\
    \textbf{ACIL-REG (Ours)} & \textbf{74.2} & \textbf{75.8} & \textbf{71.1} & \textbf{73.7} \\
    \hline
  \end{tabular}
\end{table}

\bibliographystyle{IEEEtran}
\bibliography{reference}

\end{document}